# Text vectorization via transformer-based language models and n-gram perplexities

Mihailo Škorić

University of Belgrade

mihailo.skoric@rgf.bg.ac.rs

## Abstract

As the probability (and thus perplexity) of a text is calculated based on the product of the probabilities of individual tokens, it may happen that one unlikely token significantly reduces the probability (i.e., increase the perplexity) of some otherwise highly probable input, while potentially representing a simple typographical error. Also, given that perplexity is a scalar value that refers to the entire input, information about the probability distribution within it is lost in the calculation (a relatively good text that has one unlikely token and another text in which each token is equally likely they can have the same perplexity value), especially for longer texts. As an alternative to scalar perplexity this research proposes a simple algorithm used to calculate vector values based on n-gram perplexities within the input. Such representations consider the previously mentioned aspects, and instead of a unique value, the relative perplexity of each text token is calculated, and these values are combined into a single vector representing the input.

## Background

With the emergence of *Big Data* near the beginning of the new millennium, it slowly became apparent that the separation of quality data and non-quality data is a necessity, and the research of the textual quality gained a significant amount of importance. Usual assessment methods in form human-based evaluation were subjective and, not to mention, quite expensive, so the automatic methods were used accordingly.

The most common form of model-based evaluation of text (in the field of modern natural language processing and language modelling) is perplexity, a measure of the pre-trained model's surprise with a provided input. The measure is defined as correlative to the probability that a model will generate that input, normalized by its length (Brown, et al., 1992), and is calculated as follows:

$$PP = \sqrt[n]{\frac{1}{P(w_1 w_2 \ldots w_n)}}$$

As the perplexity of a text is calculated based on the product of the probabilities of individual tokens, it may happen that one unlikely token significantly reduces the probability (i.e., increase the perplexity) of some otherwise highly probable input, while potentially representing a simple typographical error. Also, given that perplexity is a scalar value that refers to the entire input, information about the probability distribution within it is lost in the calculation. A relatively good text that has one unlikely token and another text in which each token is equally likely they can have the same perplexity value, especially for longer texts.

## Calculation

These *perplexity vectors* are calculated considering entire input, one (transformer-based) language model that is used to calculate the perplexities, and a sliding window of (given) fixed length *n* as follows (depicted in Figure 1):

1. The text is first divided into a series of tokens $w_1 w_2 \ldots w_N$, where each *w* represents one token, and *N* is their total number[1]. For the purposes of this experiment, we will identify one token with one word.
2. N-gram of size *n* are extracted one by one and passed forward to the language model used for processing. For example, if a window of size *n=3* is selected, the first n-gram will consist of the first three tokens, the second of the second three tokens, etc. ($w_1w_2w_3$, $w_2w_3w_4$, ... $w_{N-2}w_{N-1}w_N$), so the total number of n-grams will be *N – n + 1*, and the task of the language model is to calculate the perplexity measure for each of them individually.
3. For each token, *local perplexity* is calculated as an average perplexity of all n-grams that token was a part of, and an array of these, local perplexities represent the final perplexity vector.

---

[1] It should also be noted that the token does not necessarily have to refer to tokens from the dictionary prepared for the transformers, but can refer to other words and sub-words, phrases, or even sentences.

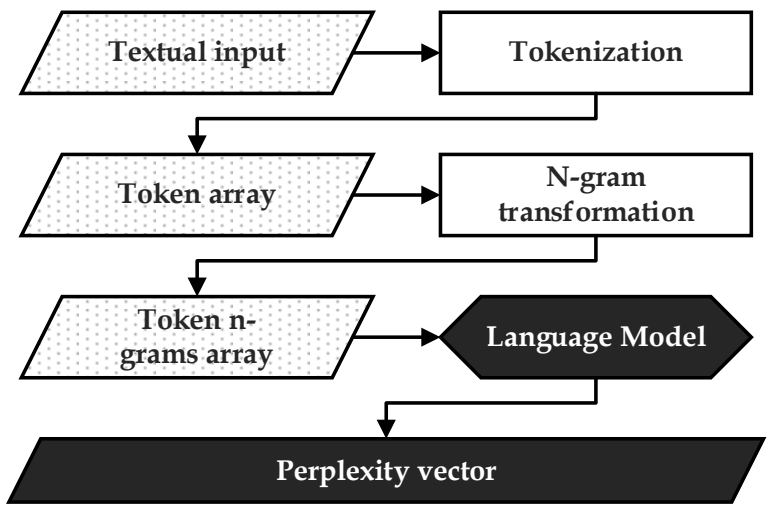

*Figure 1: Calculating the perplexity vector for an input text string using n-gram transformation and a pretrained language model.*

Let's say we have the following input string:

*When in Rome, do as the Romans do.*

Using the input and the forementioned method (for *n=4*), we first get the following list of n-grams:

1. *When in Rome,*
2. *in Rome, do*
3. *Rome, do as*
4. *, do as the*
5. *do as the Romans*
6. *as the Romans do*
7. *the Romans do.*

Then, we use a pre-trained language model (namely GPT-2 for this example) and calculate the perplexity of each text section (Table 1):

*Table 1: GPT-2* Calculated perplexity of each of the text sections created trough the n-gram (5-gram) transformation of the text.

| Text input | Perplexity |
|---|---|
| *When in Rome, do* | 76.83 |
| *in Rome, do as* | 569.06 |
| *Rome, do as the* | 111.93 |
| *, do as the Romans* | 119.84 |
| *do as the Romans do* | 72.41 |
| *as the Romans do.* | 94.20 |

We can then use this information to calculate local perplexity for each token, which we map in a new table (Table 2):

*Table 2:* Calculated local perplexity in the vicinity of each token in the input.

| Token | Local perplexity |
|---|---|
| When | 76.83 |
| in | 322.95 |
| Rome | 252.94 |
| , | 219.67 |
| do | 190.22 |
| as | 193.69 |
| the | 99.85 |
| Romans | 95.48 |
| do | 83.31 |
| . | 94.20 |

Once all the values are in place, they can be effectively visualized by mapping the values on the y-axis and the tokens (their ordinal numbers) on the x-axis (Figure 2). In this way, we get a line graph that directly shows us which parts of the text reflect the highest and which the lowest perplexity, and in so, perplexity deviation throughout the input.

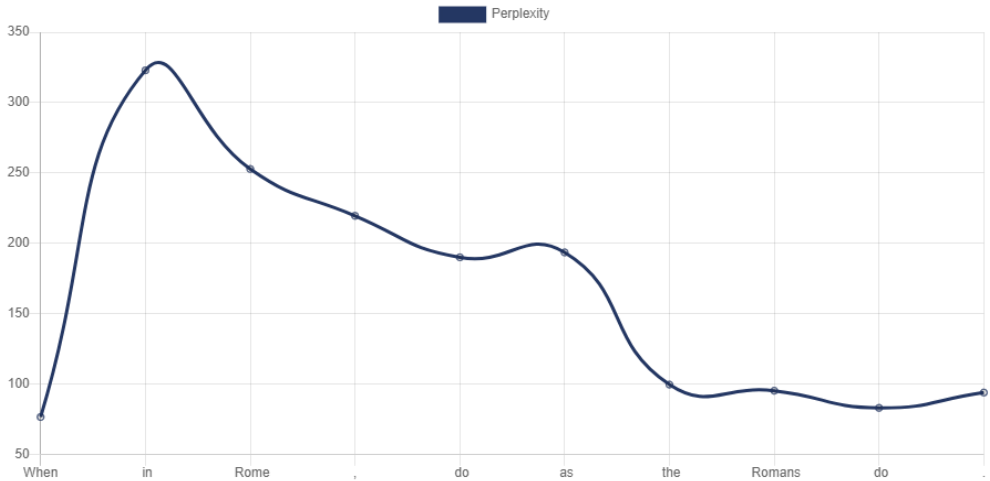

*Figure 2: The perplexity vector shown as a line graph, where the x-axis reflects the flow of the text (sentences) and the y-axis reflects the measure of perplexity.*

In addition to potentially better modeling of the perplexity of a text, this approach also enables the direct detection of words or parts of the text with the highest degree of perplexity, which may represent potentially correctable errors. From the depiction it is apparent that the second word perplexed model the most and could perhaps be corrected. For example, the curve can be flattened a bit, by using a more common wording:

*When **you are in** Rome, do as the Romans do.*

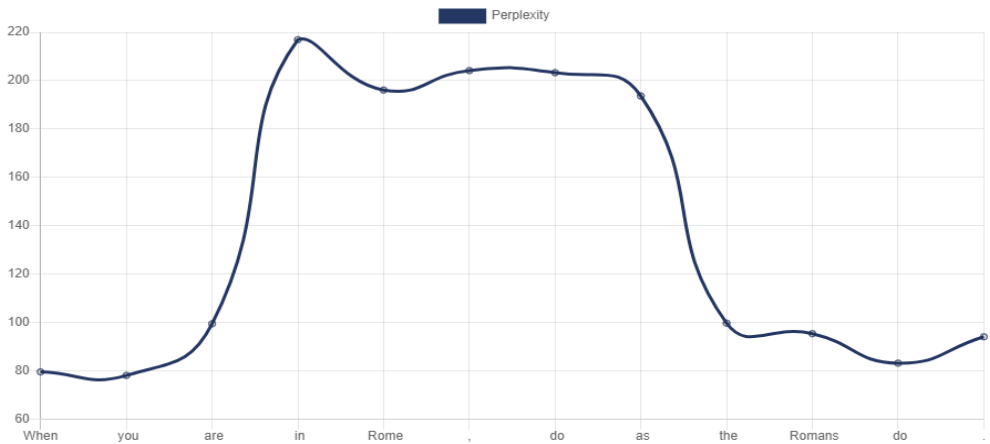

*Figure 3: The perplexity vector shown as a line graph, but for a different input.*

The graph then shows as follows (Figure 3). It is also apparent that the latter part of the graph does not change since it is outside the affective zone (it is at least five words apart from the committed changes).

## Use case

As noted in the previous section, perplexity vectors (PV) can potentially be used to find errors in text, by locating the parts of the text that have the highest relative perplexity. Hence it can be used for:

- Detection of the odd one out word in the text;
- Detection of the place in the text where the word is missing;
- Detection of the place in the text where a word was inserted by mistake.

A special evaluation dataset was prepared for each of these three examples.

Datasets for evaluating the created models are based on parallelized corpora of literary texts (literary works originally written in one of the most widespread European languages and their expert translations into the Serbian language), which were not used for training of the language models being used to calculate perplexity in order to avoid the bias during evaluation.

The first resource that was used was a fragment of the parallel Serbian-German corpus, SrpNemKor (Andonovski, et al., 2019), where only the texts of novels originally written in German were used. The second resource that was used was the parallelized translation of the third part of the Naples stories series (Perišić, et al., 2022), published within the parallel Serbian-Italian corpus created for the purposes of the *It-Sr-Ner* project, within the CLARIN organization (Krauwer & Hinrichs, 2014). A total of seven parallelized novels were used (Table 3).

*Table 3: The parallelized novels used to create the evaluation sets, their author, title, source language and number of sentences entered the corpus.*

|   | Author / Translator | Title | Word count |
|---|---|---|---|
| 1 | Tomas Bernhard / Bojana Denić | My awards | 1009 |
| 2 | Elfride Jelinek / Tijana Tropin | Pianist | 6679 |
| 3 | Milo Dor / Tomislav Bekić | Vienna, July 1999 | 1249 |
| 4 | Gunter Gras / Aleksandra Gojkov Rajić | The walk of cancer | 2868 |
| 5 | Günter de Bruyn / Aleksandra Bajazetov-Vuchen | Buridan's Donkey | 2890 |
| 6 | Christof Ransmeier / Zlatko Krasni | The Last World | 3107 |
| 7 | Elena Ferrante / Jelena Brborić | Stories about those who leave and those who stay | 8316 |

Three datasets were generated by applying simple algorithms in combination (for two out of three) with the morphological dictionary of the Serbian language (Krstev, 2008; Stanković, et al., 2018). As a prerequisite, an index i, was randomly determined for each sentence (set of expert translations), and a word with that index is selected (word in that position in the sentence). Further processing is done in accordance with the extracted word and set of sentences that we want to get.

When creating the first set (a set of chipped sentences), the selected word in each sentence was simply removed (Figure 4). In the case of creating the second set (a set of injected sentences), a new, random word, i.e. the inflected form of the word from the morphological dictionary of the Serbian language, was inserted before the selected word (at its index)(Figure 5). For the purposes of creating the last set from this group, (set of modified sentences), the selected word was replaced by another word of the same grammatical category from the morphological dictionary: e.g., an animate masculine noun in the locative singular, is replaced by another word with the same grammatical properties (Figure 6).

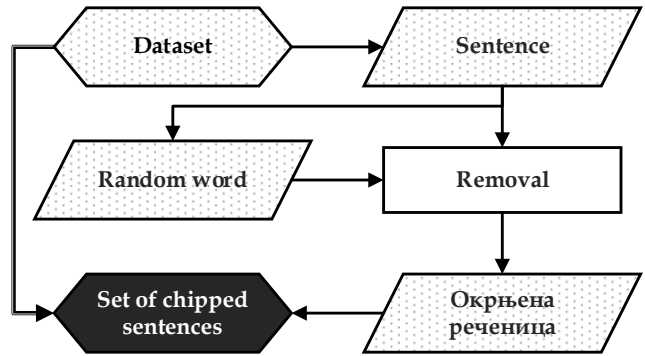

*Figure 4: Creating a set of chipped sentences from a set of expert translations.*

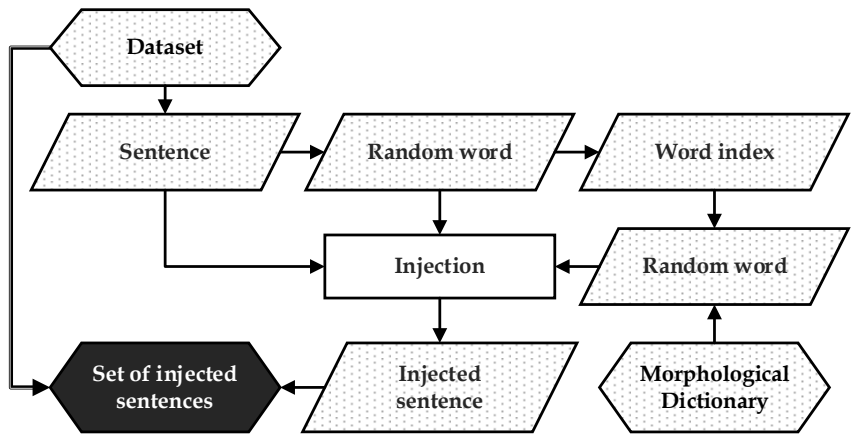

*Figure 5: Creation of a set of injected sentences from a set of expert translations by inserting a random word or inflectional form of a word from the morphological dictionary.*

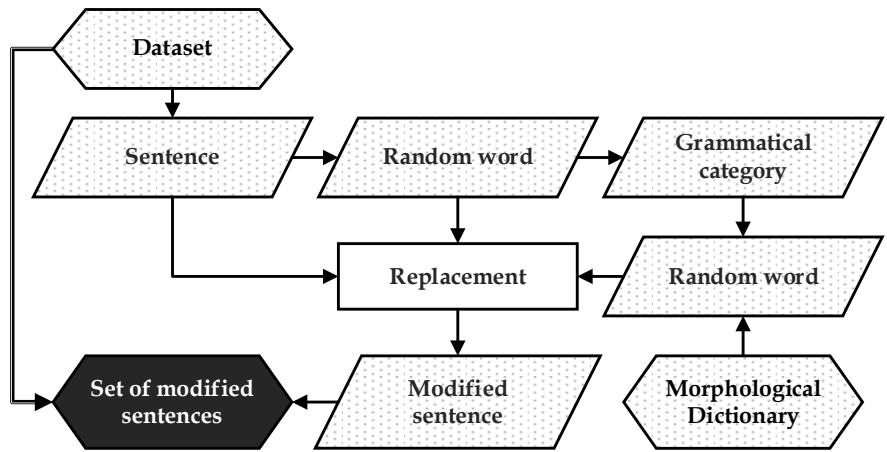

*Figure 6: Creation of a set of modified sentences from a set of expert translations by replacing a certain word with another appropriate one from the morphological dictionary.*

By applying these transformations to, for example, the given index 7 and the sentence *I remember as if it were today.* we obtain the following three sentences:

- *I remember as if it were.*
- *I remember as if it were mass today.*
- *I remember as if it flourishly.*

For evaluation, only sentences longer than seven words were used, which is two times longer than the default window used in vector creation (*n=3*). There was a total of 8188 test sentences.

During evaluation on the task of detecting removed, inserted and replaced words, perplexity vectors obtained through the processing of sentences from the prepared sets were used. Each sentence was processed using a publicly available language model for Serbian[2]. For each sentence, the odd index is selected as the index with the lowest probability measure of the perplexity vector. The accuracy of guessing the correct index was measured, with each hit affecting an increase in the accuracy measure, which was calculated as:

$$accuracy = \frac{1}{n}\sum_{i=1}^{n} \begin{cases} 0, & a_i \neq b_i \\ 1, & a_i = b_i \end{cases}$$

where *n* is the total number of sentences longer than seven words (8188), *a* is the list of indices with the highest perplexity for the vector of each of those sentences and *b* is list of indices on which each of those sentences was modified. In addition, as a basis, the method of random selection of the index for each evaluation sentence was used.

As an alternative, due to the fact that it is not as easy to guess the indices in sentences of different lengths, a measure of weighted accuracy (in relation to the sentence length) was also calculated, where each guess was counted as the difference between the number 2 and the reciprocal of the length of the sentence, so that a guess on a sentence of length one would be worth 1, (while only hypothetical, since only sentences longer than seven words are used), and hits on longer sentences were worth more than that:

$$weigthed\ accuracy = \frac{1}{n}\sum_{i=1}^{n} \begin{cases} 0, & a_i \neq b_i \\ 2 - 1/l, & a_i = b_i \end{cases}$$

---

[2] https://huggingface.co/procesaur/gpt2-srlat

where *n* is the total number of sentences longer than seven words (8188), *a* is the list of indices with the highest perplexity for the vector of each of those sentences, *b* is the list of indices on which each of those sentences was modified, and *l* is the sentence length.

The results of the experiment are shown in the table below (Table 4).

*Table 4: Evaluation results on the task of detecting the place in a sentence where a word was removed, where a word was inserted, or where one word was randomly replaced by another from the dictionary against the random selection.*

|  | set 1 | set 2 | set 3 | set 1 | set 2 | set 3 |
|---|---|---|---|---|---|---|
|  | accuracy | | | weigthed accuracy | | |
| random | 0.0580 | 0.0312 | 0.0202 | 0.1114 | 0.0600 | 0.0387 |
| calculated | 0.1037 | 0.1726 | 0.1856 | 0.2000 | 0.3339 | 0.3593 |

First of all, it should be noted that the accuracy and weighted accuracy results show a high correlation (over 99%) in the form of the Pearson correlation coefficient:

$$r = \frac{\sum_{i=1}^{n}(x_i - \bar{x})(y_i - \bar{y})}{\sqrt{\sum_{i=1}^{n}(x_i - \bar{x})^2 \sum_{i=1}^{n}(y_i - \bar{y})^2}}$$

where *n* is the size of the sample (array), *x* and *y* are the population values (accuracy and normalized accuracy), $\bar{x}$ and $\bar{y}$ are the arithmetic means of those populations, and $x_i$ and $y_i$ are the elements of the array.

From the results shown, it can be seen that the method greatly outperforms the results of random selection (with an accuracy increase of up to 827%). Also, it is apparent that it is easiest to detect the replaced word (18.56% accuracy), followed the inserted one (17.26% accuracy), while the most difficult task is to detect the removed word (10.37% accuracy).

## Conclusion

The paper describes a novel methodology in text vectorization, based on the series of n-gram perplexities calculated using a pre-trained language model (with the method being agnostic to the specific model type). The evaluation was performed on datasets of expert translations to Serbian language (which were modified in order to produce artificial mistakes) on the task of detecting the place in a sentence where a word was removed, where a word was inserted, or where one word was randomly replaced by another from the dictionary. The results indicate the superiority of the method (at least against the baseline of random selection), but the methodology requires further investigation to fully research the pros and cons.